\documentclass[conference]{IEEEtran}
\IEEEoverridecommandlockouts
\usepackage{cite}
\usepackage{amsmath,amssymb,amsfonts}
\usepackage{graphicx}
\usepackage{textcomp}
\usepackage{xcolor}
\usepackage{url}
\usepackage{algorithm}
\usepackage{algpseudocode}
\usepackage{subcaption}
\newtheorem{definition}{Definition}

\newtheorem{assumption}{Assumption}
\newtheorem{proposition}{Proposition}
\newtheorem{remark}{Remark}

\DeclareMathOperator*{\argmax}{arg\,max}
\DeclareMathOperator*{\argmin}{arg\,min}

\usepackage{mathtools}
\usepackage{breqn}

\def\BibTeX{{\rm B\kern-.05em{\sc i\kern-.025em b}\kern-.08em
    T\kern-.1667em\lower.7ex\hbox{E}\kern-.125emX}}
\begin{document}

\title{A Simple Approach to Constraint-Aware Imitation Learning with Application to Autonomous Racing}

\author{Shengfan Cao, Eunhyek Joa, Francesco Borrelli \thanks{Shengfan Cao and Francesco Borrelli are with the Department of Mechanical Engineering, University of California at Berkeley, CA 94701 USA. Eunhyek Joa is with Zoox. The work is sponsored by the Department of the Navy, Office of Naval Research ONR N00014-24-2099.  The content of the information does not necessarily reflect the position or the policy of the Government, and no official endorsement should be inferred. }}

\maketitle

\begin{abstract}
Guaranteeing constraint satisfaction is challenging in imitation learning (IL), particularly in tasks that require operating near a system's handling limits. Traditional IL methods, such as Behavior Cloning (BC), often struggle to enforce constraints, leading to suboptimal performance in high-precision tasks. In this paper, we present a simple approach to incorporating safety into the IL objective. Through simulations, we empirically validate our approach on an autonomous racing task with both full-state and image feedback, demonstrating improved constraint satisfaction and greater consistency in task performance compared to BC.  
\end{abstract}


\section{Introduction} 
Autonomous racing competitions, including the Indy Autonomous Challenge, F1TENTH, and the DARPA Grand Challenge, drive innovation in high-speed control and real-time decision-making \cite{av-racing-survey}. 
Traditional control strategies focus on optimal race-line generation but require extensive engineering and prior knowledge. 
Reinforcement learning (RL) can outperform human drivers in drone \cite{kaufmann_rl_drone} and simulated vehicle racing \cite{nature-rl-car}, yet demands risky, large-scale training and struggles with sim-to-real transfer. 
Moreover, state-of-the-art platforms often rely on advanced sensors (e.g., LiDAR, D-GPS) and significant onboard computation, limiting deployment on low-cost systems. 
Consequently, developing efficient, learning-based control for resource-constrained hardware and limited sensing remains challenging.

Imitation Learning (IL), which has demonstrated success in robotics, autonomous vehicles, and gaming \cite{imitation-learning-survey-robotics,imitation-learning-survey-av,imitation-learning-survey-gaming}, emerges as a promising candidate to address these constraints.
Recent advances in deep learning have further enhanced IL performance, allowing agents to mimic complex human strategies through the use of high-capacity models that map raw sensor data (e.g., images) to actions \cite{pan_end2end_il, loquercio_deep_drone_acrobatics}.

Despite these successes, IL-based methods often overlook safety-critical aspects of decision-making. 
In many real-world scenarios, such as autonomous driving and drone navigation, the consequences of safety constraint violations can be severe. 
Purely maximizing imitation accuracy can lead to suboptimal or unsafe maneuvers, 
especially if the agent model has insufficient representation power to perfectly clone the expert demonstrations in all possible states. 
Consequently, ensuring that an IL agent respects safety constraints (e.g., collision avoidance, maintaining certain system limits) remains an open and pressing challenge.

To address safety concerns, several recent efforts have attempted to augment IL with constraint satisfaction mechanisms. 
Many of these methods assume that the expert demonstrations themselves are safe or reliable,  thereby focusing on identifying and prioritizing safety-critical states in the dataset \cite{safe_gil, spencer_covariate_shift}. 
By emphasizing these states, the agent’s learned policy can allocate higher importance to correct behavior under dangerous conditions. 
SafeDAgger \cite{safe-dagger} takes a different approach by training an additional classifier to predict deviations from the expert and switching to a backup policy when necessary. 
However, these approaches become less effective if the expert demonstrations are themselves risky or even contain unsafe actions. 
This gap highlights the need for a more robust framework that can handle unsafe expert trajectories while still aiming to learn a safe policy.

Parallel research in classical control has long studied the problem of safe control through approaches such as Model Predictive Control (MPC), Control Barrier Functions (CBFs), and Hamilton-Jacobi reachability analysis \cite{ugo_lmpc, reference-governor, bansal_hj_reach, ad_cbf}. 
These methods provide rigorous guarantees about constraint satisfaction
by, for example, constructing safe sets or using invariant sets to ensure that the system remains within specified limits. 
A common strategy involves the use of safety filters or shielding layers that modify the control commands if a safety violation is imminent \cite{predictive_safety_filter}. 
Although such formal methods offer strong theoretical guarantees, 
they often scale poorly to high-dimensional, nonlinear systems. 
The computational overhead of solving online optimization problems or explicitly computing reachable sets becomes prohibitive in complex environments.

In this work, 
we address the problem of constraint-aware imitation learning under the realistic assumption that expert demonstrations may be unreliable or partially unsafe. 
Specifically, 
we aim to (1) learn a policy that mimics the behaviors from a set of expert demonstrations (some of which are failures)  while avoiding constraint violation, and (2) develop a safety mechanism in the training pipeline that can handle image-based inputs and partial state observations. 

Our main contributions are as follows: 
\begin{itemize}
    \item \textbf{Novel Implicit Safety Filter Architecture:} We propose a differentiable safety filter that can be embedded into classical imitation learning frameworks for high-dimensional, partially observed systems.  
    \item \textbf{Data-driven Approximation of Safe Set:} We introduce a simple data-driven approach to approximate the safe set for constrained systems inspired by \cite{ugo_lmpc}. 
    \item \textbf{Empirical Validation}: We 
    demonstrate the effectiveness of our approach on high-dimensional, nonlinear, image-feedback systems, highlighting improvements in constraint satisfaction and consistency in task performance compared to baseline methods. 
\end{itemize}

\section{Problem Formulation}
\subsection{System}
Consider a nonlinear, time-invariant, deterministic, discrete-time system described as: 
\begin{equation}\label{eq:sys_dynamics}
\begin{aligned}
    & x_{k+1} = f(x_k, u_k), ~y_k = h(x_k) + n_k,
\end{aligned}
\end{equation}
where $x_k$ is the state, $u_k$ is the control input, $y_k$ is the measurement, and $n_k$ is the measurement noise at time step $k$. 
The system is subject to constraints defined as: 
\begin{align} \label{eq:constraints}
x_k \in \mathcal{X}, ~u_k \in \mathcal{U}, ~\forall k \geq 0. 
\end{align}

As a shorthand notation, we denote $\mathbf{u}_{0:N-1} = \{u_0, \dots, u_{N-1}\}$ as a control sequence.
The state at time $N$ resulting from applying the control sequence $\mathbf{u}_{0:N-1}$ 
to the system with dynamics $f$, initialized at the state $x_0$, 
is denoted by $x_N = f(x_0, \mathbf{u}_{0:N-1})$, where each intermediate state is determined recursively with \eqref{eq:sys_dynamics}. 
We also denote a closed-loop trajecotry by $\mathbf{x}_{0:N} = \{x_0, \dots, x_N\}$. 
This implies that there exists a feasible control sequence $\mathbf{u}_{0:N-1}$ such that $x_{k+1}=f(x_k, u_k)$ for all $k = 0, \dots, N - 1$.

While a formal observability analysis is not included in this work, we assume the system is observable.  


\subsection{Constrained Optimal Control Task}
In this paper, we consider the following constrained optimal control problem. 
\begin{equation}
\label{eq:control_task_formulation}
\begin{aligned}
    \min_{\mathbf{u}_{0:\infty}} ~~ & \sum_{k=0}^{\infty} c(x_k, u_k) \\
    \textnormal{s.t.,} ~~ & x_{k+1} = f(x_k, u_k), ~ y_k = h(x_k) + n_k, \\
    & x_k \in \mathcal{X}, ~u_k \in \mathcal{U}, 
    ~~\forall k \geq 0, 
\end{aligned}
\end{equation}
where $c(\cdot, \cdot)$ is a  cost function. We assume the existence of a zero-cost target set $\mathcal{X}_f$: once the system reaches $\mathcal{X}_f$, it can remain in the set at no further cost. Specifically, we use the following assumption on $c(\cdot, \cdot)$ and $\mathcal{X}_f$.

\vspace{0.3\baselineskip}
\begin{assumption} \label{assum: forward invariant set}
    (Non-negative Cost Function and Target Set) The cost function $c(x, u)$ is non-negative for all $x\in \mathcal{X}, u\in \mathcal{U}$.
    Moreover, $\forall x \in \mathcal{X}_f$, $\exists u \in \mathcal{U}$, s.t., $f(x, u) \in \mathcal{X}_f$, $c(x,u) = 0$.
\end{assumption} 

\vspace{0.3\baselineskip}
We determine whether a trajectory is successful by checking whether its terminal state reaches the target set, namely,

\vspace{0.3\baselineskip}
\definition{\label{def:successful-failed-trajectories}
A closed-loop trajectory $\mathbf{x}_{0:N}$ is called \textbf{successful} if $x_N \in \mathcal{X}_f$, and \textbf{failed} if $x_N \notin \mathcal{X}_f$.
}
 

\subsection{Safe Imitation Learning Problem}

We are focused on the case where the controller only has access to the output $y_k$
rather than direct access to $x_k$.
Our objective is to design an output-feedback controller $\pi_\theta(y)$ that approximately solve \eqref{eq:control_task_formulation}. Assume we have access to a high-performing full-state feedback policy, denoted as $\pi_\beta(x)$, which acts as the expert.
Our approach is to mimic $\pi_\beta(x)$ for designing the output feedback controller $\pi_\theta(y)$.
Formally, this objective can be expressed as: 

\begin{equation}  
\label{eq:ideal optimization problem}
\begin{aligned}
    \min_{\theta} ~~ & 
    \mathbb{E}_{(x, y) \sim P((x, y) \mid \theta)} \left[\mathcal{L}(\pi_\theta(y), \pi_\beta(x))\right], \\
    \textnormal{s.t.} ~~ & f(x, \pi_\theta(y)) \in \mathcal{R}_\infty^B(\mathcal{X}_f),
\end{aligned}
\end{equation}
where $P((x, y)|\theta)$ denotes the distribution of state-output pairs induced by $\pi_\theta$, $\mathcal{L}(\cdot, \cdot)$ is a measurement of the discrepancy between two actions, and $\mathcal{R}_\infty^B(\mathcal{X}_f)$ is a backward reachable tube from the target set $\mathcal{X}_f$, which is defined as follows.

\vspace{0.3\baselineskip}
\definition[Backward Reachable Tube]{The N-step backward reachable tube $\mathcal{R}_N^B(\mathcal{S})$ is a set of states $x_0 \in \mathcal{X}$ that can be driven into a set $\mathcal{S}$ in $N$ time steps without constraint violation. 
Formally, 
\begin{equation} \label{eq:backward_reachable_tube_def}
    \begin{aligned}
    &\mathcal{R}_N^B(\mathcal{S}) = \\
    &~~ \left\{x_0 \mid \exists \mathbf{u}_{0:N-1}, \mathrm{s.t.}\ 
    \begin{matrix}
    f(x_0, \mathbf{u}_{0:N-1}) \in \mathcal{S}, \\
    f(x_0, \mathbf{u}_{0:k}) \in \mathcal{X}, ~\forall k < N, \\
    u_k \in \mathcal{U}, ~\forall k < N.
    \end{matrix}
    \right\}. 
\end{aligned}
\end{equation}

If $N \to \infty$, we refer to $\mathcal{R}_\infty^B$ as the infinite-time backward reachable tube. 
\label{def:backward_reachable_tube}
}
\remark{
Per Assumption \ref{assum: forward invariant set}, $\mathcal{R}_{N}^B(\mathcal{X}_f) \subseteq \mathcal{R}_{\infty}^B(\mathcal{X}_f)$, $\forall N$. 
}

\vspace{0.3\baselineskip}
\definition[Safety]\label{def:safety}{
We will refer to states in the infinite-time backward reachable tube as \textbf{safe} states, as there exists some policy that drives the system into the target set from those states. 
}

\vspace{0.3\baselineskip}
Deriving the backward reachable tube $\mathcal{R}_\infty^B(\mathcal{X}_f)$ is non-trivial, which poses a challenge in solving \eqref{eq:ideal optimization problem}. 
In practice, the safety constraint is often neglected by assuming the cloning can be sufficiently accurate and the expert is reliable and robust \cite{il-e2e-av-survey}. 
However, in applications such as autonomous racing, achieving high performance requires policies to push systems to limits. 
Any deviation by the learned policy $\pi_\theta$ from expert demonstrations, especially in unsafe directions, risks losing recursive feasibility and causing constraint violations.
Thus, prioritizing a safe operating policy over 
perfect-expert
cloning becomes essential.

In the following section, we describe how previous research tackles this challenge and their limitations. 
Then, in section \ref{sec:proposed_approach}, 
we present our approach to solving \eqref{eq:ideal optimization problem}.

\section{Related Work}
\subsection{Behavior Cloning}


One naive objective of behavior cloning is to find $\theta$ that minimizes the discrepancy between its actions and those of the expert in $\ell_2$ distance. 
\begin{align}
    \theta^{\star}_{\text{naive}} = \argmin_{\theta}\mathbb{E}_{(x, y) \sim P((x, y) \mid \theta)} \underbrace{\left[\Vert \pi_\theta(y) - \pi_\beta(x) \Vert^2\right]}_{\mathcal{L}_{\text{clone}}}. 
    \label{eq:naive_imitation_loss}
\end{align}

Let the imitation learning policy be denoted as: 
\begin{equation}
    \pi_{\text{IL}}(\cdot) = \pi_{\theta^{\star}_{\text{naive}}}(\cdot).
\end{equation}

In applications like self-driving cars \cite{bojarski-end-to-end-self-driving} and manipulation \cite{liu-bc-manipulation}, many variations of behavior cloning use \eqref{eq:naive_imitation_loss} as the fundamental building block, and add additional tunable loss terms to achieve better performance on specific tasks. 


Behavior cloning is effective in mimicking expert demonstrations in situations well-represented in its training data, but usually generalizes poorly to novel states, which is known as covariate shift \cite{il-e2e-av-survey}. 
Dataset Aggregation (\textsc{DAgger}) \cite{dagger} is an effective interactive online learning framework that mitigates this problem by collecting on-policy rollouts and use the expert to relabel the data set. 
However, DAgger's on-policy training requires costly hardware data collection. 
In addition, the theoretical zero regret promise is only achieved asymptotically, necessitating many epochs for accurate cloning. 
Models with limited representational power may not achieve zero loss even theoretically.

Another significant shortcoming of \eqref{eq:naive_imitation_loss} is its ignorance of constraints and lack of direction-specific penalties. 
Prior to achieving perfect behavior cloning (if it is even feasible), it is essential to ensure that the discrepancy is less likely in unsafe directions.

\subsection{Formal Methods for Constraint Satisfaction}
To achieve constraint satisfaction, 
common approaches in the control community include safety filters and reference governors \cite{safety-filter-survey, reference-governor}.

Safety filters are typically additional modules designed independently from the controller, and are placed between the controller and the system to project unsafe actions into a pre-designed safe set. 
This allows a high-performing policy learned within safety-ignorant frameworks like reinforcement learning to operate safely on a physical system \cite{predictive_safety_filter}. 

Let $\hat{u}_k = \pi_{\text{IL}}(y_k)$ be an output-feedback policy trained to minimize the naive $\ell_2$ imitation loss in \eqref{eq:naive_imitation_loss}, which produces potentially unsafe actions. 
We want to apply a safety filter $\pi_{\text{SF}}$
after $\pi_{\text{IL}}$ to enforce constraint satisfaction. 
A straightforward implementation is the minimum-effort predictive safety filter \cite{predictive_safety_filter}, which iteratively solves the problem in \eqref{eq:min_effort_safety_filter} and applies the first input in the optimal action sequence. 
\begin{equation}  \label{eq:min_effort_safety_filter}
\begin{aligned}
    & \min_{\mathbf{u}_{0:N-1 \mid k}} ~\Vert u_{0 \mid k} - \hat{u}_k \Vert_2^2, \\
    & ~~~~\textnormal{s.t.,} ~~~ x_{i+1 \mid k} = f(x_{i \mid k}, u_{i \mid k}), \\
    & ~~~~~~~~~~~x_{0 \mid k} = x_k, ~ x_{i \mid k} \in \mathcal{X}, ~ u_{i \mid k} \in \mathcal{U}, \\
    &  ~~~~~~~~~~~x_{N \mid k} \in \mathcal{R}_\infty^B(\mathcal{X}_f), ~~~\forall i=0, \dots, N-1, 
\end{aligned}
\end{equation}
where $N$ is the horizon of the predictive safety filter and $u_{i|k}$ and $x_{i|k}$ are the input and the predicted state at time step $k+i$, respectively. 
After solving \eqref{eq:min_effort_safety_filter}, the optimal safety filter policy
\begin{align}
    u_k = \pi_{\text{SF}}(\hat{u}_k \mid x_k) = u^{\star}_{0\mid k} 
\end{align}
is applied to the system \eqref{eq:sys_dynamics}.
Note that \eqref{eq:min_effort_safety_filter} relies on full-state feedback.


Other variations of safety filters are formulated based on formal methods such as CBF \cite{cbf-e2e-safety-filter, cbf-safety-filter}, Hamilton-Jacobi reachability analysis \cite{bansal_hj_reach}, but their overarching goal is similar to \eqref{eq:min_effort_safety_filter}.
Reference governors \cite{reference-governor}, on the other hand, approach the problem by proactively modifying the reference signal before it reaches the controller to ensure the system closely follows the reference without violating the constraints. 


However, most existing formal methods for constraint satisfaction assume that full-state information is available or can be estimated accurately. 
Many real-world applications operate primarily on high-dimensional sensor data (e.g., camera images), and extending safety filters to operate directly in an image-based or partial-observation setting is an outstanding challenge.

\section{Proposed Approach: \\ Constraint-aware Behavior Cloning}\label{sec:proposed_approach}

In this section, we present a constraint-aware imitation learning framework to approximately solve \eqref{eq:ideal optimization problem}. Given an expert policy $\pi_\beta(x)$, we incorporate the safety filter \eqref{eq:min_effort_safety_filter} into behavior cloning \eqref{eq:naive_imitation_loss}, formulating output-feedback policies $\pi_\theta(y)$ that leverage the filter as an additional training-time expert.
Specifically, our goal is to directly recover 
\begin{align}
\theta^{\star}_{\text{CA}} = \argmin_{\theta}  
    \mathbb{E}_{(x, y) \sim P((x, y) \mid \theta)} \underbrace{\left[\mathcal{L}(\pi_\theta(y), \pi_{\text{SF}}(\pi_{\text{IL}}(y) \mid x))\right]}_{\mathcal{L}_{\text{CA}}}
    \label{eq:CABC-general-objective}
\end{align}
from end-to-end training with an privileged expert that has full state feedback, which aims to achieve both high performance and constraint satisfaction. 
In the scope of this paper, we use the naive behavior cloning objective in \eqref{eq:naive_imitation_loss} and the min-effort predictive safety filter in \eqref{eq:min_effort_safety_filter} as an example to facilitate analysis. 

\subsection{Construction of $\mathcal{L}_{\text{CA}}$}
\label{sec:constraint-aware-bc}
In this section, we discuss the construction of $\mathcal{L}_{\text{CA}}$ in \eqref{eq:CABC-general-objective}. 
As a first step, we approximately reformulate \eqref{eq:min_effort_safety_filter} to make it differentiable, enabling its use in backpropagation.

Let $N=1$ for the safety filter formulated in \eqref{eq:min_effort_safety_filter}.
With Definition \ref{def:backward_reachable_tube}, 
the constraints in \eqref{eq:min_effort_safety_filter} can be written as:
\begin{align}
    f(x_k, u_{0 \mid k}) &\in \mathcal{R}_\infty^B(\mathcal{X}_f), ~~u_{0 \mid k} \in \mathcal{U}. \label{eq:full_state_feedback_mpc_backward_reachable_tube_version_constraints}
\end{align}
The minimum effort safety filter $\pi_{\text{SF}}$ in \eqref{eq:min_effort_safety_filter} can be approximated by $\pi_\xi$ by softening the constraint. 
\begin{equation}
\begin{aligned}
    \pi_\xi(\hat{u}_k \mid x_k) \triangleq \argmin_{u_{0 \mid k} \in \mathcal{U}} &\Big[\Vert u_{0 \mid k} - \hat{u}_k \Vert^2 \\ &+ \mathbb{I}_{\mathcal{R}_\infty^B(\mathcal{X}_f)}(f(x_k, u_{0 \mid k}))\Big], \label{eq:min_effort_safety_filter_unconstrained}
\end{aligned}
\end{equation}
where 
$\mathbb{I}$ is implemented as a soft indicator function 
\begin{align}
    \mathbb{I}_{\mathcal{R}_\infty^B(\mathcal{X}_f)}(x) &= -\lambda \log p(x \in \mathcal{R}_\infty^B(\mathcal{X}_f)), \label{eq:nll_as_indicator_fn}
\end{align}
and $\lambda > 0$ is a hyperparameter. 
Note that if the classifier $p$ is a hard classifier that always correctly outputs 0 for unsafe or 1 for safe, \eqref{eq:nll_as_indicator_fn} strictly enforces safety. 

Suppose $p(x \in \mathcal{R}_\infty^B(\mathcal{X}_f))$ is given as a differentiable function. Then,  $\pi_\xi(\hat{u}_k \mid x_k)$ in \eqref{eq:min_effort_safety_filter_unconstrained}-\eqref{eq:nll_as_indicator_fn} can be numerically approximated with gradient-based methods. 

Next, we seek to integrate \eqref{eq:naive_imitation_loss} and \eqref{eq:min_effort_safety_filter_unconstrained} into a joint objective $\mathcal{L}_{\text{CA}}$. In this step, we use the following assumption.

\vspace{0.3\baselineskip}
\assumption[Zero-bias training of $\pi_{\text{IL}}$]{
To facilitate derivation of $\mathcal{L}_{\text{CA}}$, 
we assume the naive behavior cloning policy $\pi_{\text{IL}}$ clones the expert policy $\pi_\beta$ with no bias on the training set $\mathcal{D}$, i.e., $\mathbb{E}_{(x, y) \sim \mathcal{D}} [\pi_{\text{IL}}(y) - \pi_\beta(x)] = 0$. 
Additionally, assume the dataset $\mathcal{D}$ has no covariate shift. 
}

\vspace{0.3\baselineskip}
Plug $\hat{u}_k = \pi_{\text{IL}}(y)$ and \eqref{eq:min_effort_safety_filter_unconstrained} into \eqref{eq:CABC-general-objective},
\begin{align}
\theta^{\star}_{\text{CA}} 
    =& \argmin_{\theta} \mathbb{E}_{(x, y) \sim P((x, y) \mid \theta)} \Big[\Vert \pi_\theta(y) - \pi_{\text{IL}}(y) \Vert_2^2 \notag \\ 
    & \qquad - \lambda \log p\left(f(x, \pi_\theta(y)) \in \mathcal{R}_\infty^B(\mathcal{X}_f)\right)\Big] \notag \\
    =& \argmin_{\theta} \mathbb{E}_{(x, y) \sim P((x, y) \mid \theta)} \Big[ \underbrace{\Vert \pi_\theta(y) - \pi_\beta(x) \Vert_2^2}_{\mathcal{L}_{\text{clone}}}  \notag \\
    & \qquad \underbrace{- \lambda \log p\left(f(x, \pi_\theta(y)) \in \mathcal{R}_\infty^B(\mathcal{X}_f)\right)}_{\mathcal{L}_{\text{safety}}}\Big]. 
    \label{eq:constraint_aware_bc_obj}
\end{align}  

\eqref{eq:constraint_aware_bc_obj} is the joint objective $\mathcal{L}_{\text{CA}}$ of the naive behavior cloning agent $\pi_{\text{IL}}$ and the safety filter $\pi_{\text{SF}}$.
Note that \eqref{eq:constraint_aware_bc_obj} contains the naive behavior cloning loss $\mathcal{L}_{\text{clone}}$, same as in \eqref{eq:naive_imitation_loss}, and an additional NLL loss $\mathcal{L}_{\mathrm{safety}}$. 
This can be considered as concurrently learning from two experts, one that optimizes performance and one that enforces constraint satisfaction. 
We refer to the expert providing $\mathcal{L}_{\mathrm{safety}}$ as the \textbf{safety critic}. 

\subsection{Tractable Reformulation of $\mathcal{L}_{\mathrm{safety}}$} \label{sec:safety-auto-labeling}
The safety critic consists of two components: the forward dynamics $f(x, u)$, and the safety likelihood $p(x \in \mathcal{R}_\infty^B(\mathcal{X}_f))$, 
both assumed to be unknown.

We use the learned function approximators $\hat{f}_{\phi_f}(x, u)$ and $\hat{p}_{\phi_p}(x)$ as their corresponding surrogates. 
Specifically, the forward dynamics estimator $\hat{f}_{\phi_f}$ is trained via auto-regression, i.e., 
\begin{align}
    \phi_f^{\star} &= \argmin_{\phi_f} \mathbb{E}_{(x, u, x') \sim \mathcal{D}} [ \Vert x' - \hat{f}_{\phi_f}(x, u) \Vert^2], 
    \label{eq:autoregression_mseloss}
\end{align}
\text{where} $x^\prime=f(x, u)$. 

The safety likelihood estimator $\hat{p}_{\phi_p}$ 
is learned using the binary cross-entropy loss, 
i.e., 
\begin{multline}
    \phi_p^{\star} = \argmax_{\phi_p} \mathbb{E}_{x \sim \mathcal{D}} \Bigg[ 
    \Big( s \log \hat{p}_{\phi_p}(x)  + \\
    (1 - s) \log (1 - \hat{p}_{\phi_p}(x)) \Big) 
    \Bigg],
    \label{eq:safety_clf_bceloss}
\end{multline}
where
\begin{align}
    s = \left\{\begin{matrix}
        1, & x \in \mathcal{R}_\infty^B(\mathcal{X}_f), \\
        0, & x \notin \mathcal{R}_\infty^B(\mathcal{X}_f).
    \end{matrix}\right.
\end{align}

However, computing $s$ is challenging due to unknown $\mathcal{R}_\infty^B(\mathcal{X}_f)$
, which is difficult to compute. \footnote{For low-dimensional linear systems, the backward reachable tube can be computed numerically \cite{cinf-set-propagation}. 
The safety of individual states can also be approximated using the feasibility of an MPC controller with perfect modeling of the dynamics and the constraints \cite{predictive_safety_filter}.
However, 
it is generally a challenge to compute $\mathcal{R}_\infty^B(\mathcal{X}_f)$ for a black-box, high-dimensional, nonlinear system.} 
Next we present a simple self-supervised  
approach to acquire surrogate labels $s$ to learn $\hat{p}_{\phi_p}$ from closed-loop trajectories without any knowledge of the system, called \textbf{safety auto-labeling}.

Let $\mathcal{D} = \{\mathbf{x}_{0:T_j}^{(j)}\}$ be a dataset collected by rolling out a data collection policy $\pi_{\mathrm{collect}}$ in closed-loop from various initial states, where $j$ is the index of the trajectory in the dataset. 
Assume $\pi_{\mathrm{collect}}$ is designed such that it can generate both successful and failed iterations. 

Recall Definition \ref{def:successful-failed-trajectories} for successful and failed trajectories. 
The dataset $\mathcal{D}$ can be partitioned into $\mathcal{D}_+ = \{\mathbf{x}_{0:T_j}^{(j)} \mid x_{T_j}^{(j)} \in \mathcal{X}_f\}$ and $\mathcal{D}_? = \{\mathbf{x}_{0:T_j}^{(j)} \mid x_{T_j}^{(j)} \notin \mathcal{X}_f\}$. 

By Definition \ref{def:backward_reachable_tube} and \ref{def:safety}, all states visited during successful iterations are safe, i.e., 
\begin{align} \label{eq:D+_are_safe}
    \mathcal{D}_+ \subseteq \mathcal{R}_\infty^B(\mathcal{X}_f).
\end{align}

As its counterpart, ideally we need $\mathcal{D}_- \subseteq \overline{\mathcal{R}_\infty^B(\mathcal{X}_f)}$ to construct the loss in \eqref{eq:safety_clf_bceloss}. 
However, states in $\mathcal{D}_?$, visited during failed iterations, are not necessarily unsafe. 

\vspace{0.3\baselineskip}
\proposition{\label{thm:local_cvx_overapprox}
Suppose $\mathcal{R}$ is a set of interest, and we have a set of samples $\mathcal{S}$ from $\mathcal{R}$. 
Let $B_\rho(x)$ be a ball with radius $\rho$, centered at $x$, and $\text{conv}(\cdot)$ be the convex hull of a set. 
If $x \in \text{conv}(\mathcal{S} \bigcap B_\rho(x))$, then $d(x, \partial \mathcal{R}) \geq -\rho$, where \footnote{$\partial \mathcal{R}$ is the boundary of the set $\mathcal{R}$.} 
$$d(x, \partial \mathcal{R}) = \left\{\begin{matrix}
    \inf_{y \in \partial \mathcal{R}} \Vert x - y \Vert, ~\text{if}~x \in \mathcal{R}, \\
    -\inf_{y \in \partial \mathcal{R}} \Vert x - y \Vert, ~\text{if}~x \notin \mathcal{R}.
\end{matrix}\right.$$ 
If, in addition, $\mathcal{R}$ is locally convex at $x$ within radius $\rho$, then $d(x, \partial \mathcal{R}) \geq 0$. 
}

Let
\begin{align}
    D_- \triangleq \mathcal{D}_? \setminus \{x \in \mathcal{D}_? \mid x \in \text{conv}(\mathcal{D}_+ \bigcap B_\rho(x))\}. 
\end{align}
Proposition \ref{thm:local_cvx_overapprox} indicates that by excluding points that are in the convex hull of states known to be safe, $\mathcal{D}_-$ contains only states that are likely unsafe with higher confidence, and thereby reducing false negatives in the binary labels. 



\begin{figure}[htbp]
    \centering
    \includegraphics[width=0.95\linewidth]{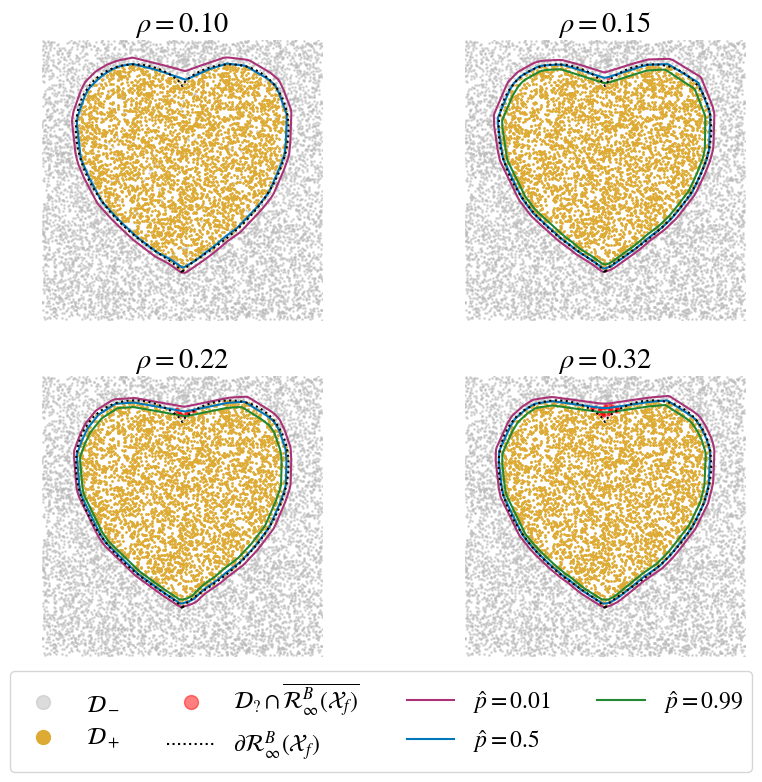}
    \caption{
    Illustration of the proposed safety auto-labeling algorithm and the corresponding decision boundary in a constructed example. The \textbf{dotted black curve} is the true boundary of $\mathcal{R}_\infty^B(\mathcal{X}_f)$. $\rho$'s are the radius of the nearest neighbor in Alg.~\ref{alg:Constraint-aware-IL}. $\mathcal{D}_+$ is uniformly sampled from $\mathcal{R}_\infty^B(\mathcal{X}_f)$, and $\mathcal{D}_?$ is uniformly sampled from $\mathcal{X} \supset \mathcal{X}_f$. The \textbf{blue curve} is the decision boundary of an MLP as a binary classifier trained to classify states in $\mathcal{D}_+$ and $\mathcal{D}_-$, and the \textbf{purple and green curves} are its level curves of higher confidence. \textbf{Red points} are states incorrectly filtered from $\mathcal{D}_?$. Note that as $\rho$ decreases, less points are incorrectly removed from $\mathcal{D}_?$ in the concave regions, resulting in a better estimation of the true boundary in those regions. 
    }
    \label{fig:self-labeling_example}
\end{figure}

Figure~\ref{fig:self-labeling_example} illustrates an example of the proposed auto-labeling technique and its effect on binary safety classification. 
With $\mathcal{D}_+$ sufficiently covering $\mathcal{R}_\infty^B(\mathcal{X}_f)$, and $\mathcal{D}_?$ sufficiently covering $\mathcal{X}$, the decision boundary $\hat{p} = 0.5$ closely aligns with its true boundary, particularly when $\rho$ is appropriately small.
As long as the set remains locally convex within a radius of $\rho$, the method correctly identifies points inside $\mathcal{R}_\infty^B(\mathcal{X}_f)$ from $\mathcal{D}_?$. 

Without assuming the local convexity property, the choice of $\rho$ is critical. Note that in the example, the true set boundary is concave with a sharp corner. 
In an online iterative learning framework, $\rho$ should adapt to the sample density of $\mathcal{D}_+$, decreasing as density increases. A large $\rho$ may over-smooth concave boundaries, while a small $\rho$ risks false negatives for not getting sufficient neighbors for comparisons, affecting classification stability. 
Although we do not dynamically modify $\rho$ during training in our experiments, Figure~\ref{fig:self-labeling_example} captures the core intuition that as sample density of $\mathcal{D}_+$ increases, dynamically decreasing $\rho$ enables $\hat{p}$ to first capture the macroscopic structure of $\mathcal{R}_\infty^B(\mathcal{X}_f)$ before refining finer details. 

Note that the proposed local convex hull-based safety boundary estimation can be extended to account for different types of uncertainty by incorporating robustified convex hull approximations.


\subsection{Proposed learning framework}

The proposed learning framework is described in Algorithm \ref{alg:Constraint-aware-IL}. 
This framework builds upon DAgger by incorporating the training of $\hat{f}_{\phi_f}$ and $\hat{p}_{\phi_p}$ into the process, and incorporating the necessary auto-labeling process for their training. 


\begin{algorithm}[htbp]
\caption{Constraint-aware Behavior Cloning}\label{alg:Constraint-aware-IL}
\begin{algorithmic}
\State Initialize $\mathcal{D}_+$, $\mathcal{D}_?$ as empty sets 
\State Appropriately initialize $\theta$, $\phi_f$, $\phi_p$
\For{$j \gets 0, \dots, M - 1$} 
    \State $\pi_{\textnormal{collect}} \gets \alpha^j \pi_\beta + (1 - \alpha^j) \pi_\theta$ \Comment{$\alpha$ is a hyperparameter}
    \State Rollout $\pi_{\textnormal{collect}}$ to collect $\mathcal{D}_{\textrm{new}}$
    \State Partition $\mathcal{D}_{\text{new}}$ into $\mathcal{D}_+^{\prime}$ and $\mathcal{D}_?^{\prime}$
    \State $\mathcal{D}_+ \gets \mathcal{D}_+ \bigcup \mathcal{D}_+^{'}$, $\mathcal{D}_? \gets \mathcal{D}_? \bigcup \mathcal{D}_?^{\prime}$
    \For{$x \in \mathcal{D}_?$} \Comment{ Safety auto-labeling.}
    \State $\mathcal{S} \gets \text{RadiusNeighbors}(x \mid \mathcal{D}_+, \rho)$
    \If{$x \in \text{GetConvexHull}(\mathcal{S})$} 
        \State Remove $x$ from $\mathcal{D}_?$
        \Comment{Remove if likely safe.}
    \EndIf
    \EndFor
    \State 
    $\textnormal{ComputeGradient}(\mathcal{D}_+, \mathcal{D}_? \mid \pi_\theta, \hat{f}_{\phi_f}, \hat{p}_{\phi_p})$ \Comment{See Fig.~\ref{fig:training_diagram}}
    \If{$j \bmod k_f = 0$} \Comment{$k_f = 5$ in our impl.}
    \State $\phi_f \gets \textnormal{Update}(\phi_f \mid \nabla_{\phi_f} L_{\textnormal{dyn.}})$ 
    \EndIf
    \If{$j \bmod k_p = 0$} \Comment{$k_p = 10$ in our impl.}
    \State $\phi_p \gets \textnormal{Update}(\phi_p \mid \nabla_{\phi_p} L_{\textnormal{critic}})$
    \EndIf
    \State $\theta \gets \textnormal{Update}(\theta \mid \nabla_\theta L_{\textnormal{agent}})$
\EndFor
\end{algorithmic}
\end{algorithm}

\begin{figure*}[htbp]
    \centering
    \includegraphics[width=\linewidth]{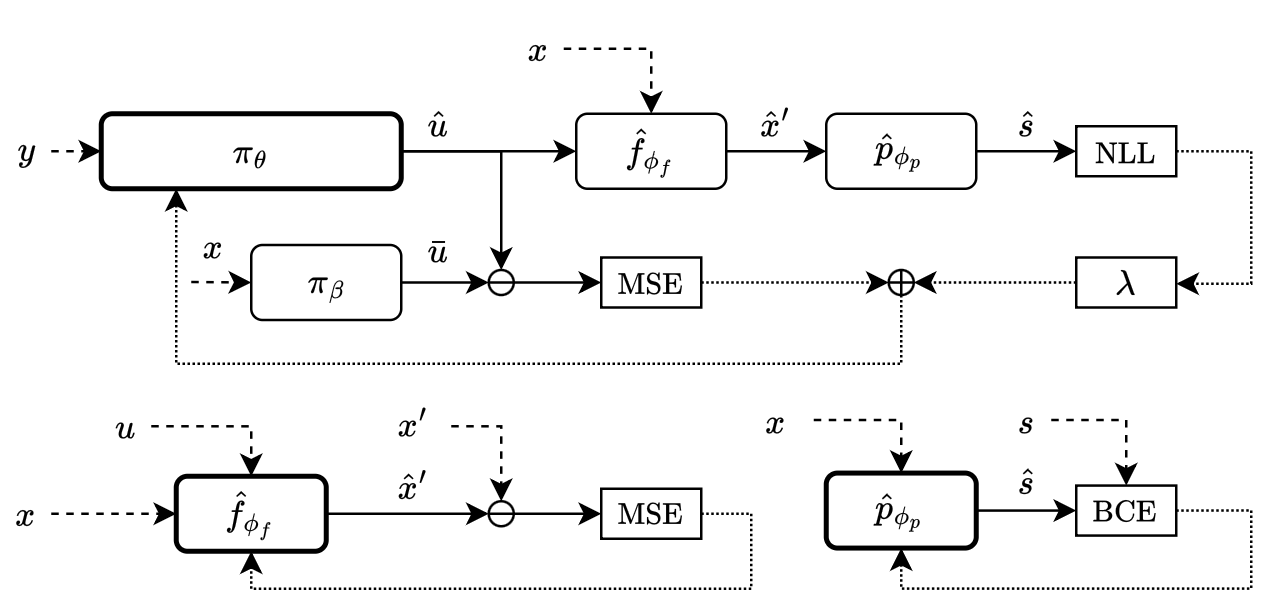}
    \caption{
    Gradient Computation in the Proposed Architecture.
    \textbf{Dashed lines} represent inputs from the dataset $\mathcal{D}$, where each entry consists of state $x$, system output $y$, safety label $s$, closed-loop action $u$, and next state $x'$. 
    \textbf{Dotted lines} indicate the correspondence between each loss function and its target module.
    \textbf{Bold-faced blocks} indicate trainable modules, while non-bold blocks remain frozen during training.
    \textbf{Rectangular blocks} represent loss functions: MSE (mean squared error), NLL (negative log-likelihood), and BCE (binary cross-entropy).
    During test time, only the policy network $\pi_\theta$ is used, while all other modules participate solely in training.
    }
    \label{fig:training_diagram}
\end{figure*}

\section{Experiments}
This section presents the empirical results of the proposed learning framework. 
We apply our approach in a CARLA-based autonomous racing simulation \cite{carla} with customized vehicle dynamics. 
The objective is to finish 50 consecutive laps with minimum lap time while avoiding collision with the boundary of the track. \footnote{An implementation of the
experiments in this paper can be found at \url{https://github.
com/CadenzaCoda/ConstraintAwareIL.git}.}

The state $x$ in the experiments is modeled as $x = \begin{bmatrix}
    v_{\text{long}} & v_{\text{tran}} & \omega_\psi & s & x_{\text{tran}} & e_\psi
\end{bmatrix}^T$, 
where $v_{\text{long}}$, $v_{\text{tran}}$, and $\omega_\psi$ are the longitudinal velocity, lateral velocity and the yaw rate; $s$ is the arc length along the reference path; $x_{\text{tran}}$ is the lateral deviation, and $e_\psi$ is the heading error. 
These are defined in the Frenet frame, which moves along the reference path with the longitudinal axis aligned with the path tangent and the lateral axis normal to it. 

The inputs are $u = \begin{bmatrix}
    u_{\text{a}} & u_{\text{steer}}
\end{bmatrix}^T$, corresponding to the throttle and steering control of the car. 
The output $y$ consists of an RGB image from a front-facing camera and velocity measurements, i.e., $y = \begin{bmatrix}
    \text{Img}(x) & v_{\text{long}} & v_{\text{tran}} & \omega_\psi
\end{bmatrix}$. 
These output feedback signals are assumed to be available because they can also be readily obtained on a physical platform equipped with a camera, an IMU, and wheel encoders.

\begin{figure}[htbp]
    \centering
    \begin{subfigure}[b]{0.45\linewidth}
        \centering
        \includegraphics[width = \linewidth]{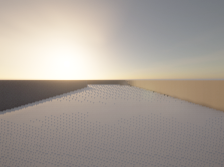}
        \caption{}
    \end{subfigure}
    \begin{subfigure}[b]{0.45\linewidth}
        \centering
        \includegraphics[width = \linewidth]{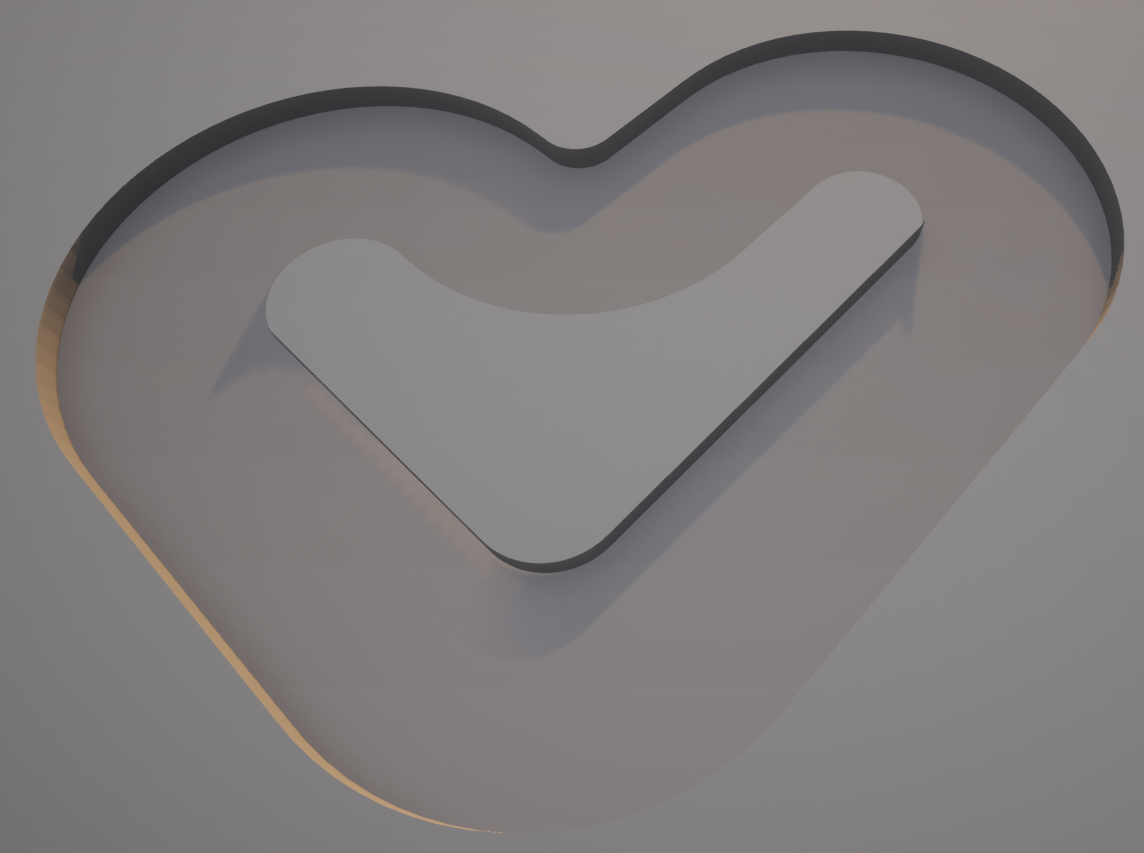}
        \caption{}
    \end{subfigure}
    \caption{The CARLA-based simulation environment for the experiments. \textbf{Left:} Example of onboard front-facing camera view $\text{Img}(x)$. \textbf{Right:} Geometry of the race track.}
    \label{fig:image-based-exp-env}
\end{figure}

Fig.~\ref{fig:image-based-exp-env} shows an example of the first-person camera view and the top view of the race track. 
Note that the direction of the sunlight is randomized with each environment reset. 
We further assume the image contains no distinguishable landmark to uniquely localize the ego vehicle. 
The ground truth vehicle dynamics in the simulator 
is treated as a black box non-linear system in the training pipeline. 

We utilize the \textbf{early stopping (ES)} technique to stop the training when the evaluation performance is at its peak. 
This is a common technique in machine learning to avoid overfitting to the measurement noise, causing the performance to degrade. 
The early stopping condition here is when the controller completes 50 consecutive laps for the second time. 

Also, as a baseline for our approach in all experiments, we choose the naive behavior cloning objective in \eqref{eq:naive_imitation_loss} and the DAgger to train a policy with the same architecture.

\subsection{Image Feedback Autonomous Path Following} 
\label{sec:image-feedback-path-following}
We first apply our method to learn to follow the track at a conservative speed. 
The expert $\pi_\beta(x)$ is a PID controller, tuned to track the center line at 1 m/s. 
The policy $\pi_\theta(y)$ has access to image and velocity measurements. 
The architecture of $\pi_\theta(y)$ is based on ResNet18, with the final linear layer replaced with a three-layer MLP with 128 hidden neurons. 

\begin{figure}[htbp]
    \centering
    \includegraphics[width=0.9\linewidth]{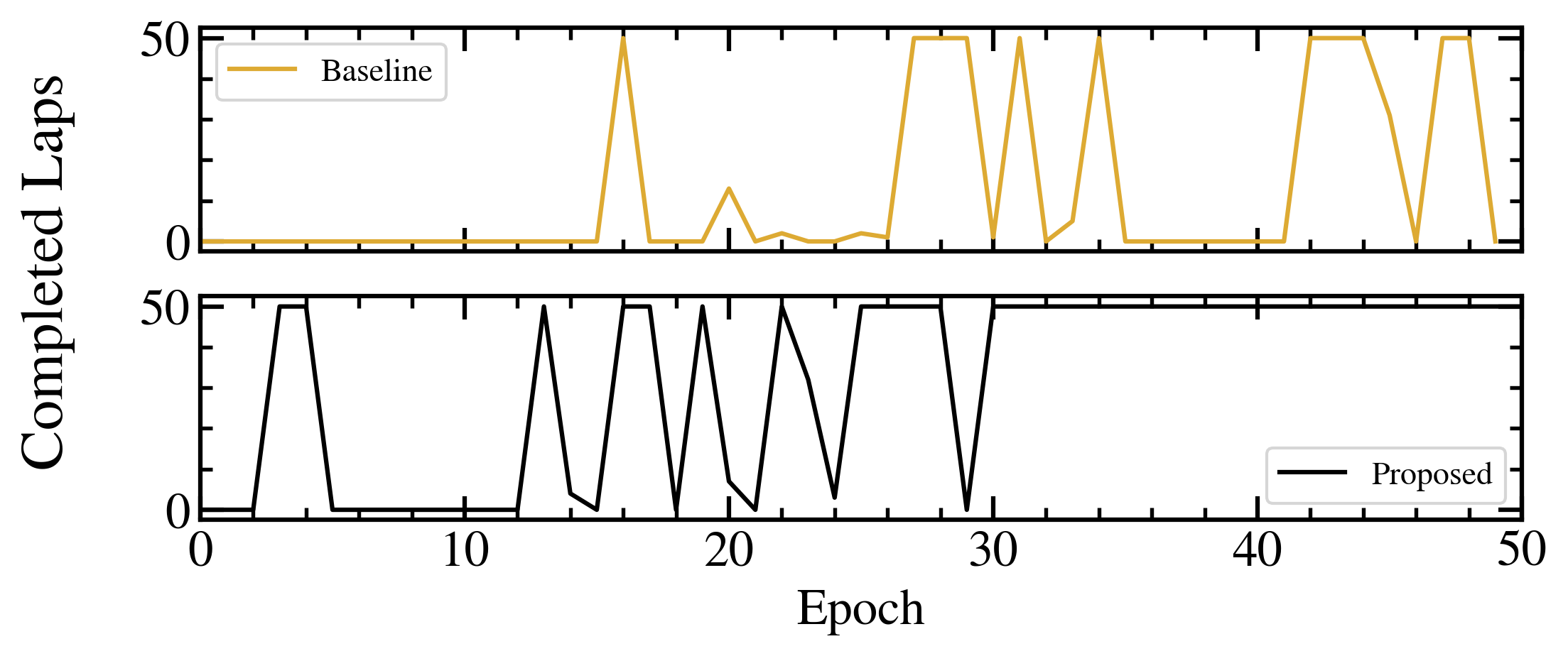}
    \caption{
    Number of laps without constraint violation for baseline vs. propose (Exp. \ref{sec:image-feedback-path-following}). 
    The x-axis is the number of training epochs and 
    the y-axis is the successful iterations completed without constraint violation. 
    The tests are initialized at the same initial condition and are truncated once the controller completes 50 laps. 
    In the proposed method, $\lambda = 1$, $\rho = 1$. 
    }
    \label{fig:image-feedback-path-following}
\end{figure}

Figure~\ref{fig:image-feedback-path-following} compares the proposed method with the baseline. 
As shown, both approaches lead to a policy that can complete 50 consecutive laps. 
However, the proposed method achieved this with fewer training epochs.

\subsection{Full-state Feedback Autonomous Car Racing} \label{sec:full-state-racing}
Next, we apply our approach to a high-speed racing task. 
In only this example, we allow the policy to observe the full state. 
The expert policy $\pi_\beta$ is an MPCC-conv controller\cite{mpcc-conv}, optimized for high performance without hard constraints to stay on track, and often operates near the system’s limits.
The architecture of policy $\pi_\theta$ in this experiment is restricted to 3-layer MLPs with 128 hidden neurons. 

\begin{figure}[htbp]
    \centering
    \includegraphics[width=\linewidth]{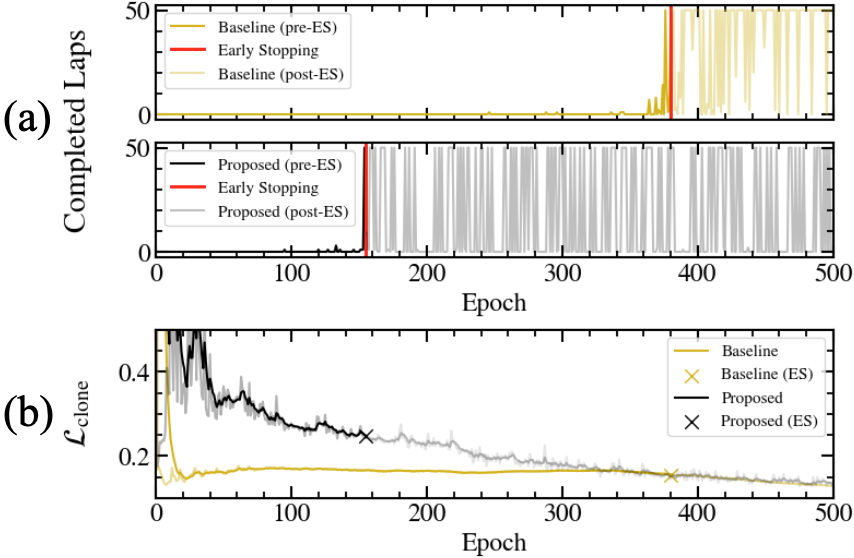}
    \caption{\textbf{(a):} Number of laps without constraint violation for baseline vs. proposed (Exp. \ref{sec:full-state-racing}). 
    The tests are initialized at the same initial condition and are truncated once the controller completes 50 laps.
    \textbf{Early stopping} is applied when a controller completes 50 laps for the second time. 
    In the proposed method, $\lambda = 10$, $\rho = 1$. 
    \textbf{(b)} Imitation loss for baseline vs.\ proposed; $\times$ marks early stopping (second epoch of 50-lap-safe policy recovery). 
    Although imitation loss converges more slowly, our approach recovers a performant policy in significantly fewer epochs, demonstrating more effective learning.
    }
    \label{fig:experiment-2}
\end{figure}

\begin{figure}[htbp]
    \centering
    \includegraphics[width=\linewidth]{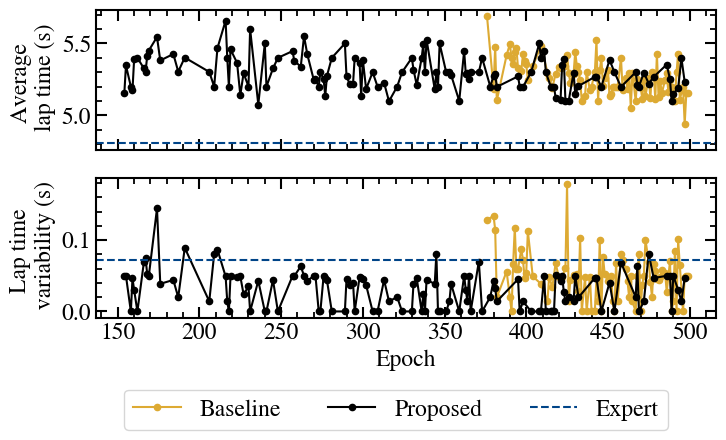}
    \caption{Lap time and average and variability of the baseline and the proposed approach in Experiment \ref{sec:full-state-racing}. The variability is characterized by the standard deviation of lap times. The plot only shows epochs where the controller completes 50 laps. 
    The dashed line shows the statistics of $\pi_\beta$. }
    \label{fig:experiment-2-eval}
\end{figure}

Fig.~\ref{fig:experiment-2} and \ref{fig:experiment-2-eval} show the comparison between the proposed and baseline method in the test time performance. 

Precision is key in this task, as we observed constraint violations can be caused by small deviations, especially when making a tight turn at high speed. 
Without a safety-specific penalty in the learning objective, it takes extreme behavior cloning precision, and consequently, many training epochs, to consistently avoid constraint violation. 
This poses a significant challenge, especially because $\pi_\beta$ is not robust to large input disturbance. 

In contrast, our approach effectively reduced the likelihood of unsafe deviation from the expert, therefore leading to higher overall return with less training effort. 
Although our approach reduces imitation loss more slowly, it attains a recursively feasible policy in far fewer epochs than the baseline. 
This shows that incorporating a safety penalty is a more efficient route to a high-performance, safe controller, especially if the expert policy is not robust to actuation noise.

We chose the number of completed laps as our performance metric to demonstrate recursive feasibility. 
We observed that in closed loop, the metric is nearly binary, either 50 laps or early termination. 
Note that the imitation loss is only a proxy for performance, which
can decrease smoothly while the realized policy remains brittle. 
Accordingly, we adopt early stopping when the controller consistently completes 50 laps.

\subsection{Image-feedback Autonomous Car Racing} \label{sec:image-feedback-racing}
Next, we explore the task with the same setup and objective described in \ref{sec:full-state-racing}, but the policy can only observe the output of the system. 
The architecture of $\pi_\theta$ is based on ResNet18, with the final linear layer replaced with a 3-layer MLP with 128 hidden neurons. 

\begin{figure*}[htbp]
    \centering
    \includegraphics[width=0.8\linewidth]{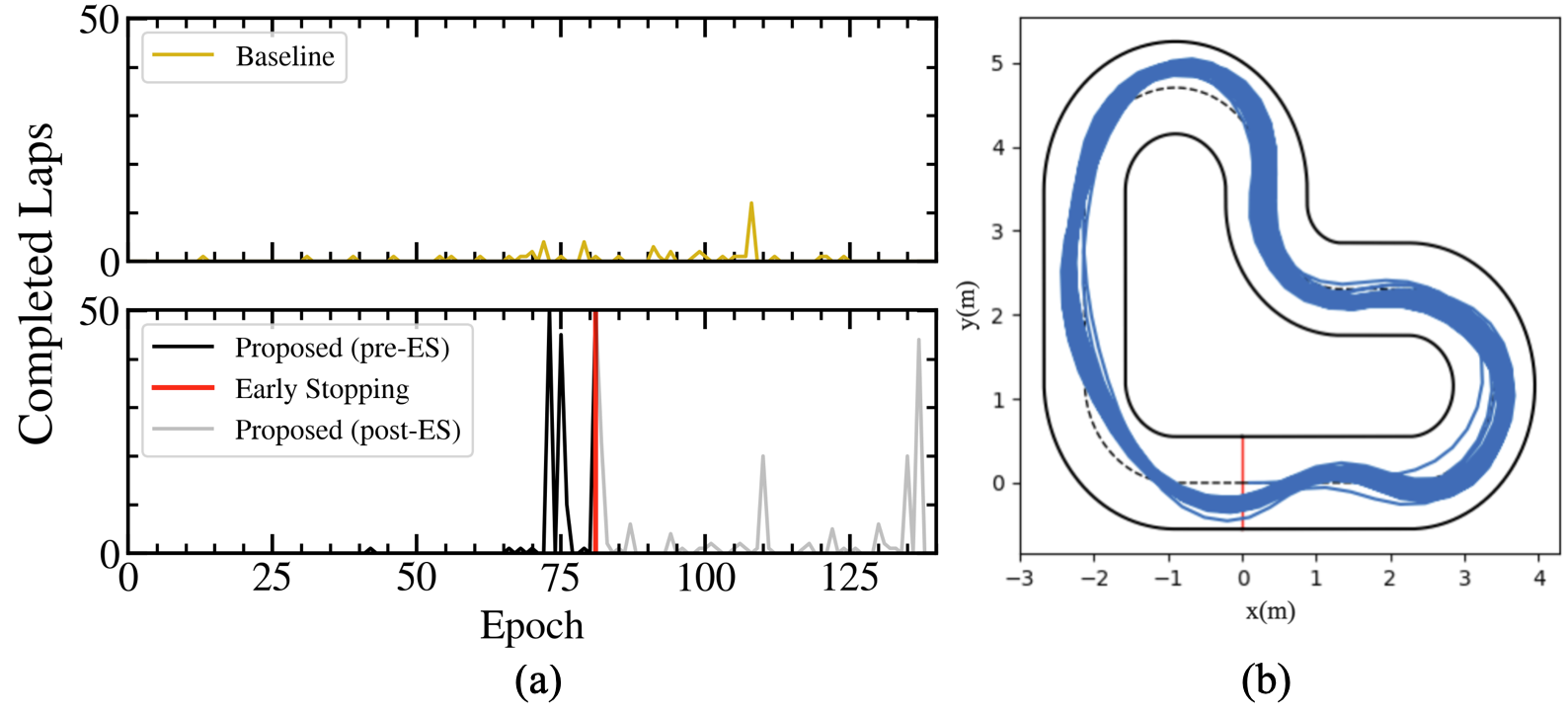}
    \caption{
    \textbf{(a):} Number of laps without constraint violation for baseline vs. propose (Exp. \ref{sec:image-feedback-racing}). 
    In the baseline, $\lambda=1$, $\rho = 1$. 
    The setting is the same as Experiment \ref{sec:full-state-racing}, except the controllers must rely only on image and velocity feedback instead of the full state. 
    In the proposed method, $\lambda = 1$, $\rho = 0.5$. 
    \textbf{Early stopping} (ES) is applied when a controller completes 50 laps for the second time. 
    \textbf{(b):} Rollout of $\pi_\theta$ with proposed method when early stopping is triggered. The policy completed 50 consecutive laps with the average lap time = $5.65$
    , max lap time = 5.8 seconds, and min lap time = 5.5 seconds. 
    In comparison, the average expert lap time = 4.805 seconds.
    }
    \label{fig:experiment-3}
\end{figure*}

Fig.\ref{fig:experiment-3}a compares the performance of the proposed method against the baseline. The proposed method successfully recovered a 50-laps-safe policy within 80 epochs, whereas the baseline failed to exceed even 10 laps. 
Fig.\ref{fig:experiment-3}b illustrates the rollout trajectory of $\pi_\theta$ when early stopping occurs at epoch 81. The trajectory indicates that $\pi_\theta$ learned to maintain a safe distance from walls while achieving consistent lap times.

The partial observability and the sensor noise made this task particularly challenging, and 
we observed particularly high variability in training dynamics in this case. 
However, we consistently noticed a spike in test time performance at around 100 epochs, and saw an overall reduced failure rate and improved consistency across 
various hyper-parameter settings. 

Considering the challenge imposed by partial observability and noise, we suggest applying early stopping when the performance is at its peak, rather than continuing training in the hope that the performance will further improve.


\section{Discussion and Future Work}
In this work, we introduced a learned safety penalty into the imitation learning objective to address safety-critical tasks. 
Our experimental results demonstrate that this addition substantially reduces constraint violation rates and stabilizes performance, particularly during the earlier stages of training. 
Moreover, we empirically showed the effectiveness of the approach in vision-based end-to-end learning tasks. 
Notably, our method also achieves safe policy recovery in tasks requiring operation near the system’s handling limits, a setting where traditional imitation learning often struggles.

For future work, we plan to extend the proposed approach to other imitation learning objectives and safety filters for improved performance. 
Additionally, we aim to address the high-variance training dynamics caused by the non-stationary optimization landscape introduced by the evolving safety penalty term. Strategies such as adaptive hyperparameter scheduling and curriculum learning may help mitigate variance and improve reliability.
Finally, while we show strong results in simulation, we expect real-world differences to degrade performance. 
We plan to mitigate this via robust or domain-randomized expansions of our training and by carefully validating $\hat{p}_{\phi_p}$ on physical systems.

\end{document}